\documentclass[10pt,twocolumn,letterpaper]{article}

\usepackage{cvpr}
\usepackage{times}
\usepackage{epsfig}
\usepackage{graphicx}
\usepackage{amsmath}
\usepackage{amssymb}

\usepackage{textcomp}

\usepackage[breaklinks=true,bookmarks=false]{hyperref}

\cvprfinalcopy 

\newcommand{\norm}[1]{\left\lVert#1\right\rVert} 

\def\cvprPaperID{****} 

\ifcvprfinal\fi
\begin{document}

\title{Curve Reconstruction via the Global Statistics of Natural Curves}

\author{Ehud Barnea and Ohad Ben-Shahar\\
	Dept. of Computer Science, Ben-Gurion University\\
	Beer-Sheva, Israel\\
	{\tt\small \{barneaeh, ben-shahar\}@cs.bgu.ac.il}
}

\maketitle

\begin{abstract}
	Reconstructing the missing parts of a curve has been the subject of much computational research, with applications in image inpainting, object synthesis, etc. Different approaches for solving that problem are typically based on processes that seek {\em visually pleasing} or {\em perceptually plausible} completions. In this work we focus on reconstructing the underlying {\em physically likely} shape by  utilizing the global statistics of natural curves. More specifically, we develop a reconstruction model that seeks the {\em mean physical curve} for a given inducer configuration. This simple model is both straightforward to compute and it is receptive to diverse additional information, but it requires enough samples for all curve configurations, a practical requirement that limits its effective utilization. To address this practical issue we explore and exploit statistical geometrical properties of natural curves, and in particular, we show that in many cases the mean curve is scale invariant and oftentimes it is extensible. This, in turn, allows to boost the number of examples and thus the robustness of the statistics and its applicability. The reconstruction results are not only more physically plausible but they also lead to important insights on the reconstruction problem, including an elegant explanation why certain inducer configurations are more likely to yield consistent perceptual completions than others.
\end{abstract}

\section{Introduction}
\label{sec:intro}

\begin{figure}
	\begin{center}
		\begin{tabular}{cc}
			\includegraphics[height=2.5cm, trim={0cm 0cm 0 0.35cm}]{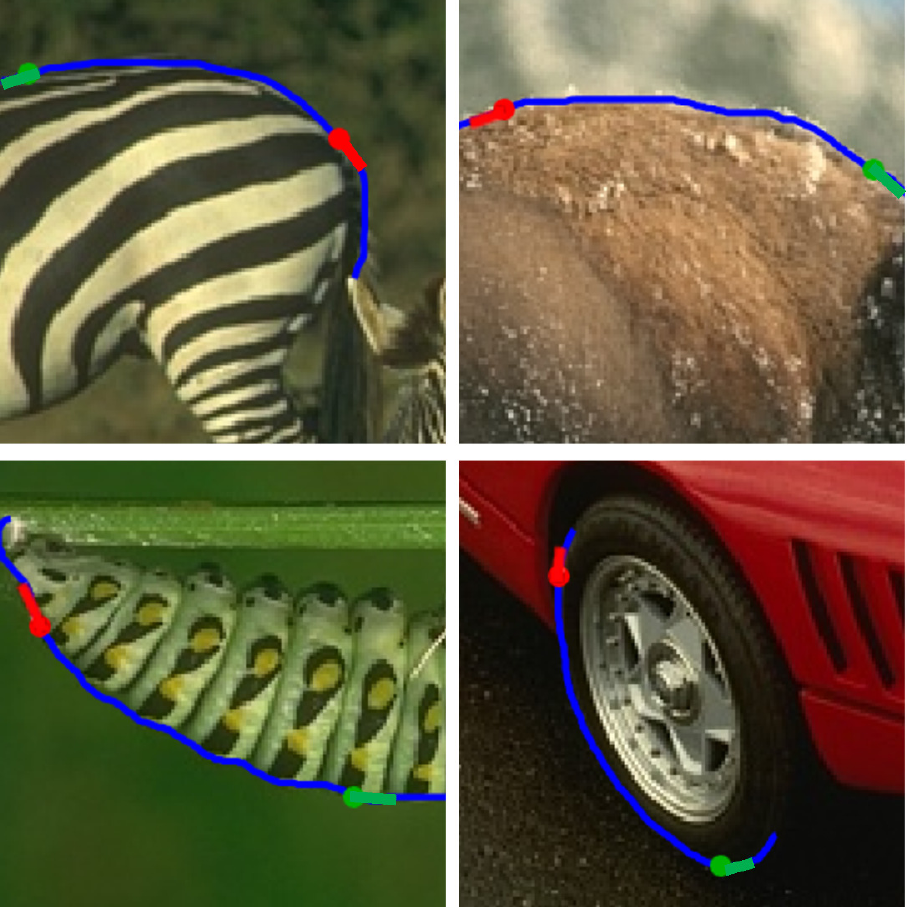} &
			\includegraphics[height=2.6cm, trim={0.7cm 0.5cm 0 0.45cm},clip]{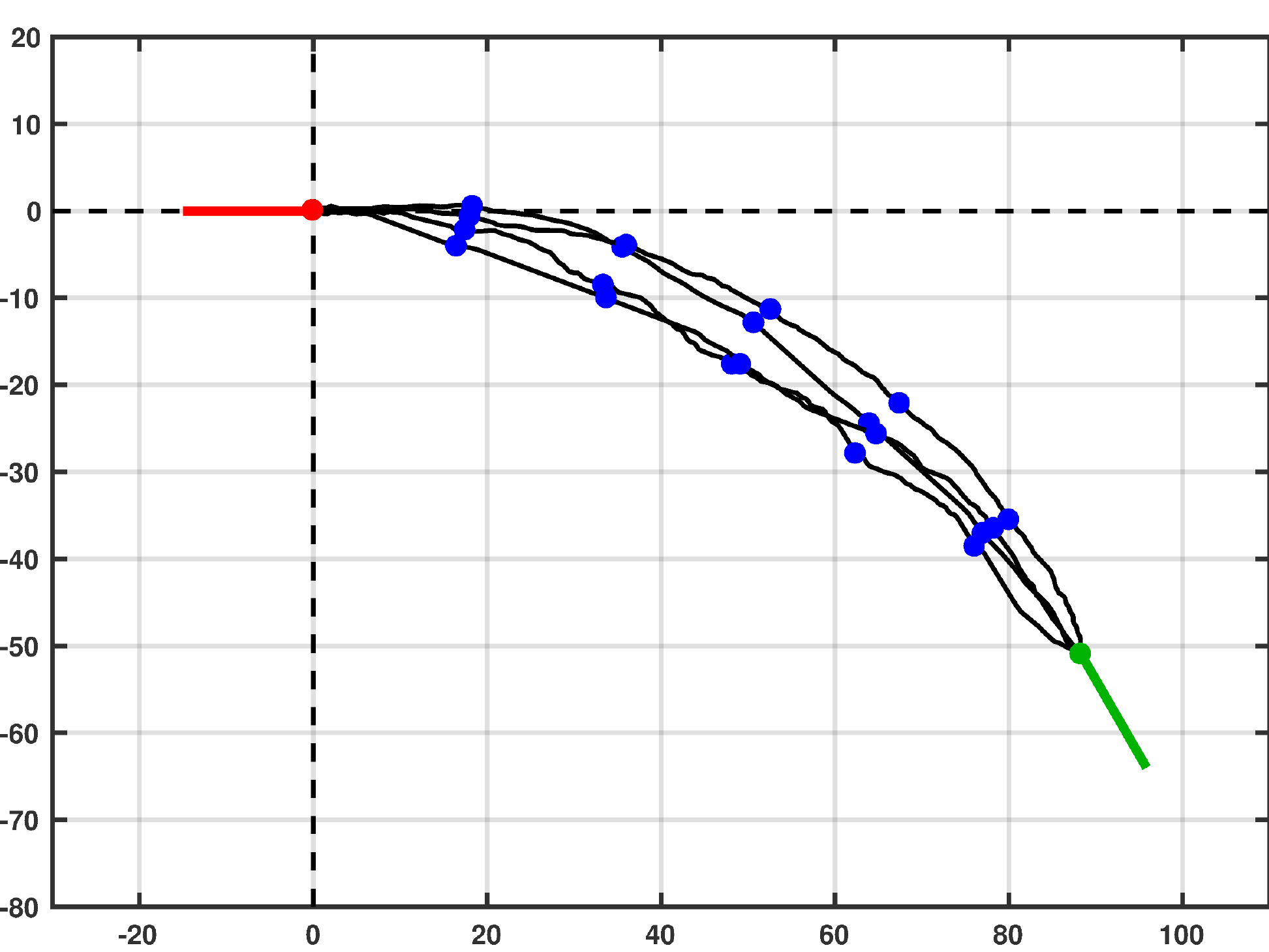} \\
			(a)&(b) \\
			\includegraphics[height=2.6cm, trim={0.8cm 0.9cm 0 0.45cm},clip]{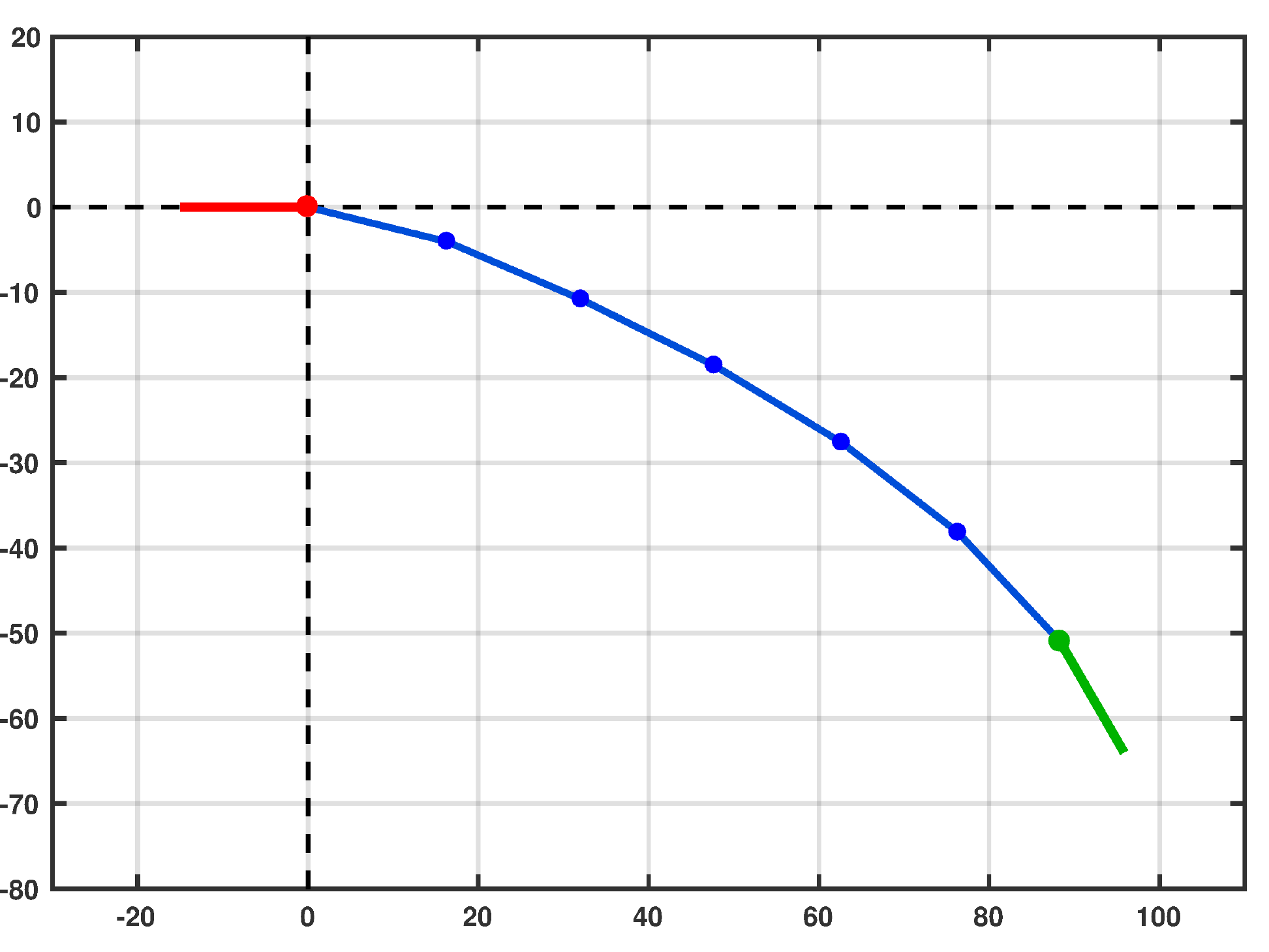} &
			\includegraphics[height=2.6cm, trim={0.7cm 0.5cm 0 0.45cm},clip]{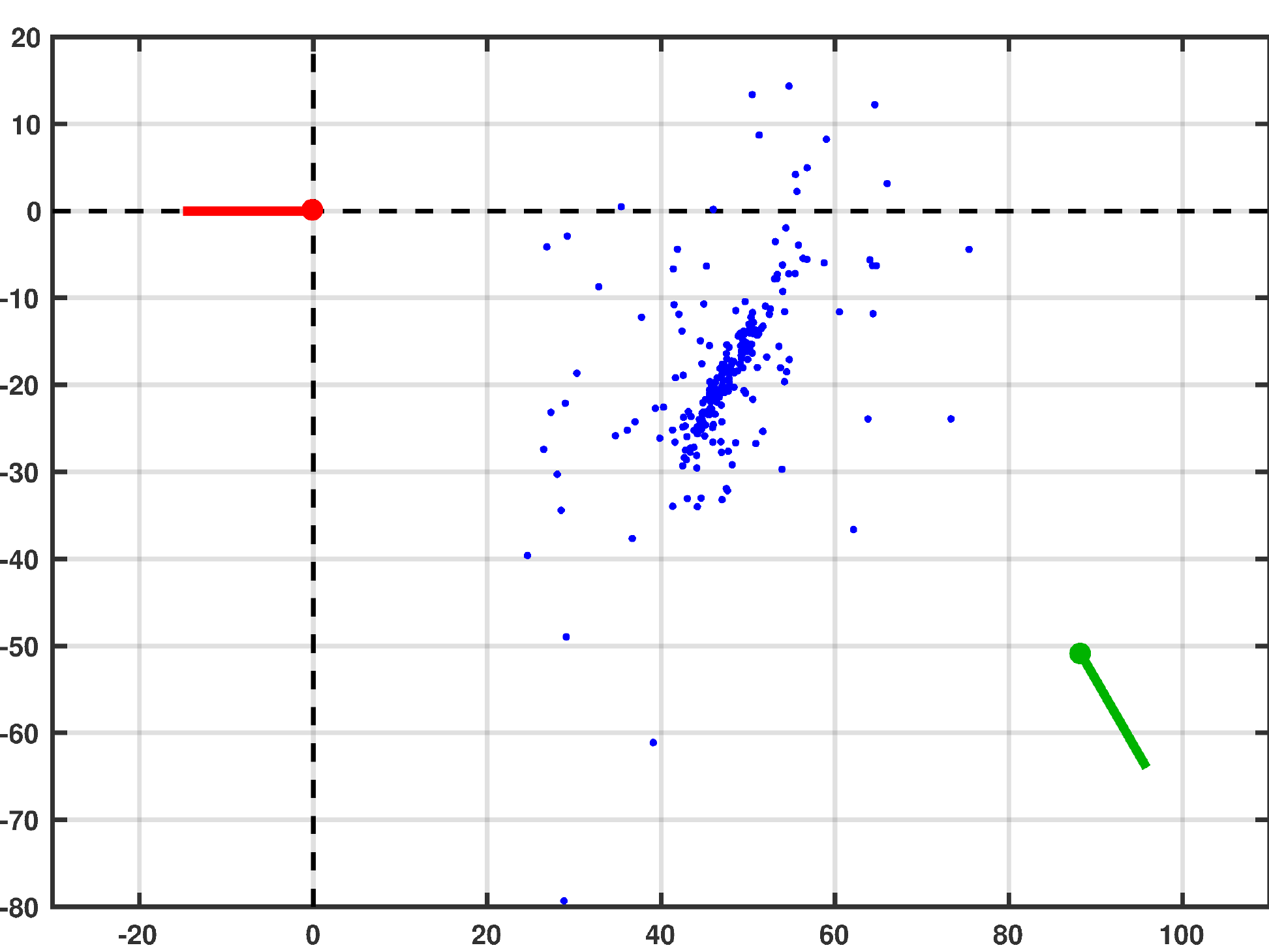} \\
			(c)&(d) \\
		\end{tabular}
	\end{center}
	\caption{Curve reconstruction via global statistics of natural curves. 
		{\bf (a)} To measure natural curve prior for a given relative inducer configuration, curve fragments
		having the desired relative inducers (end points + tangent orientation) are collected from ground truth
		curves labeled by humans~\cite{guo_kimia_2012_CVPRW}.  
		{\bf (b)} The curves are then normalized to the same frame of reference and sampled along their arc length 
		with $n$ equally spaced points. In this example $n=5$ and to avoid clutter only 4 curves are shown. 
		{\bf (c)} Each point $i$ along the arc length of the reconstructed curve  
		is calculated as the mean of the corresponding points in the dataset, or more abstractly, 
		as the expected value of the distribution of all $i^{th}$ points of curves that match the same inducers. 
		{\bf (d)} While some variance {\em is} observed among the centers of all 245 
		curves, the distribution appears rather tight, somewhat anisotropic, and approximately normal,
		suggesting that the expected value nicely represents the likely outcome.
	}
	\label{fig:summary}
	\vspace{-7pt}
\end{figure}

The reconstruction of visual curves is the process of filling in curve fragments that are completely unobservable due to occlusion or adversarial acquisition conditions.  When attempting this task one could pursue one of two different goals, either the original {\em physical} shape or the {\em perceived} one. (Note that oftentimes these two curves are quite different.). Different models for the generation of missing parts of curves were indeed suggested \cite{Ullman_1976_BC,Kimia_etal_1999_POCV,Mumford_1994_in_Algebric_Geometry_and_its_applications,Horn_1983_ATMS,Ben-Yosef_Ben-Shahar_2012_PAMI,harary_tal_2012_CompGeom,ren_malik_2002_ECCV,zhou_etal_2012_CompsAndGraphics} and employed in various applications such as image inpainting 
\cite{voronin_etal_2012_SPIE,shibata_etal_2011_ACCV,sun_etal_2005_tog,wei_liu_2016_SigImgVideoProc}, and computer graphics
\cite{harary_tal_2012_CompGeom,harary_and_tal_2011_CGF}.

With much of the theoretical work done in the context of curve {\em completion} that supports perceptual plausibility (\ie, matching the perceived completion under cases such as occlusion), most studies have focused on defining different shape criteria for the generation of visually pleasing curves (for example those with minimal total curvature \cite{Ullman_1976_BC,Mumford_1994_in_Algebric_Geometry_and_its_applications,Horn_1983_ATMS} or minimal change of curvature \cite{Kimia_etal_1999_POCV}) rather than relying on any measurable prior on natural curves. 
Additionally, the computation of these curves entails a minimization procedure that usually requires an iterative scheme that is susceptive to local minima and cannot be easily expanded to include additional information (such as the curvature at the visible end points or the shape of the occluder, among others). Critically, no consensus exists in the literature regarding which completion criteria are preferred, and evaluation is hardly performed vis-a-vis physical objects (whose shape one tries to reconstruct) or perceptual findings.

Somewhat differently, and avoiding arbitrary shape constraints as above, Ben-Yosef and Ben-Shahar suggested a biologically plausible theory based on an abstraction of the primary visual cortex and a least action criterion \cite{Ben-Yosef_Ben-Shahar_2012_PAMI}. This was recently elaborated as a constrained Elastica model in the unit tangent bundle with emergent properties like sensitivity to curvature~\cite{Ben-Shahar_Ben-Yosef_2015_PAMI}. Other works employed probabilistic schemes such as first-order and multi-scale higher-order Markov models \cite{Williams_Jacobs_1997_NC,ren_malik_2002_ECCV}. These models too, however, suffer from some of the issues discussed above. 

In this work we suggest to incorporate a stronger and more realistic prior for reconstruction by following the global shape statistics of natural curves. Like others, we will assume that the reconstruction is performed between two {\em inducers} -- a pair of end points, their tangent orientation, and possibly additional information about the observed part of the curve as it penetrates the occluder. Simply put, we first sample from the distribution of natural curves that corresponds to each relative inducer configuration. This operation can be done empirically from annotated datasets~\cite{guo_kimia_2012_CVPRW}. Representing each curve as a discrete set of $n$ points distributed uniformly along the arc length, we then extract the point-wise mean of the collected curves, from which a ``mean curve'' can be compiled. In contrast to other methods employing only \emph{local} statistics of adjacent point pairs, this computation examines curves in their entirety, thus employing \emph{global} statistics.

Even when lacking any local consideration, this process generates surprisingly smooth and visually pleasing curves, as shown in Fig.~\ref{fig:summary}. And yet, for many inducer configurations, especially for longer curves, the number of observed samples may be small, prohibiting robust statistics and thus proper reconstruction. To alleviate this problem we show that the mean curve is scale invariant in many cases and further investigate its extensibility, allowing to collect and generate many more samples for each configuration, providing a significantly more robust reconstruction process.

In addition to their pleasing appearance, the generated curves and our proposed completion model enjoy several favorable properties. By design, the resultant curves strongly represent the distribution of physical image curves and thus the structure of objects in natural images. By definition of the expected value, these curves are also  ``closest'' to all dataset curves with the same inducer configuration. 
The proposed method can also easily extend to include the curvature at the end points, the shape of the occluder, or any other information, by simply conditioning the measured probability distribution (and thus the reconstruction computation) on the desired properties or conditions.
Furthermore, the proposed reconstruction approach provides novel insights into the reconstruction problem. For example, we show that for ``convenient'' configurations (when the inducers are relatable~\cite{Kellman_Shipley_1991_Cognitive_Psychology} or even just ``facing'' each other) the distribution of curves in natural images is quite narrow and the mean curve is similar to most ground-truth curves, providing a reconstruction that closely matches the ground truth physical curves in most cases. However, curves with ``abnormal'' or ``difficult'' inducer configurations (e.g., when the inducers face away from each other) exhibit much greater variance, which implies that any reconstruction, regardless of the underlying principle employed, is likely to generate a curve that does not reproduce the original shape. Such an analysis suggests that in these cases additional information is strongly recommended, as well as a model that can exploit it (like the one suggested here).

\section{Prior art}
\label{sec:prev}

Different methods have been suggested for the generation of curves, with either the goal of completing curves when no information is known apart for the inducer configuration (usually the result of occlusions), or when the curve part to complete is visible but difficult to discern due to noisy or faint edge appearance. To the best of our knowledge no work has focused on reconstruction of missing curve parts to match the physical reality, and so our review of the prior art focuses on methods obtaining visually pleasing or perceptually plausible completions over large gaps.

Much of the previous work on curve completion is based on defining a set of axioms or properties that the completed curve should satisfy, in what has been dubbed the {\em axiomatic approach}~\cite{Ben-Yosef_Ben-Shahar_2012_PAMI, Ben-Shahar_Ben-Yosef_2015_PAMI}. Given such axioms, a model and computational approach that generate such curves is then developed.  Among the first is the Biarc model suggested by Ullman~\cite{Ullman_1976_BC} that seeks curves of least curvature that are also smooth, isotropic, and extensible. Loosening some of these constraints, the Biarc model constructs the completed curve using two circular arcs (that meet at a curvature discontinuity). Taking least curvature strictly later gave rise to the Elastica model~\cite{Mumford_1994_in_Algebric_Geometry_and_its_applications,Horn_1983_ATMS,Bruckstein_Netravali_1990_CVGIP,Sharon_etal_2000_PAMI,Weiss_1988_CVGIP,mio_etal_2006_IJCV}. A different property, suggested by Kimia \etal \cite{Kimia_etal_1999_POCV}, is the minimization of change in curvature, resulting in a family of visually pleasing curves known as Euler Spirals. This completion model has been improved later in various follow up papers~ \cite{xu_etal_2013_CVPR,zhou_etal_2012_CompsAndGraphics,walton_meek_2008_CRV}.

Employing a different approach, in what has been dubbed the ``mechanistic approach''~\cite{Ben-Yosef_Ben-Shahar_2012_PAMI, Ben-Shahar_Ben-Yosef_2015_PAMI}, some works define probabilistic models for the generation of maximum probability curves or for the calculation of point-wise probabilities of belonging to a curve. In Williams and Jacobs \cite{Williams_Jacobs_1997_NC}, for example,  the completion is described as the most probable random walk between the curve inducers. Being a first order Markov model, this work can be seen as minimizing some local probabilistic measure along the curve while ignoring long range interactions. A similar model was also employed together with local detection responses for grouping of visible edges \cite{felzenszwalb_and_McAllester_2006_CVPRW}. In a later work, it was shown that this Markov assumption does not comply with curvature statistics of natural curves, and so a higher order model was suggested incorporating multiple scales for the task of edge boundary detection \cite{ren_malik_2002_ECCV}.

Seeking to benefit the advantages of both approach types while avoiding arbitrary or unfounded assumptions about the shape of the completed curves, Ben-Yosef and Ben-Shahar~\cite{Ben-Yosef_Ben-Shahar_2012_PAMI} moved the computation framework from the image plane to a space that abstracts the primary visual cortex where the visual system allegedly performs the curve completion task.  With a proper abstraction the problem was then considered in the unit tangent bundle and the completion criteria become ones that are inspired by how biological neural circuits may behave. The simple models that emerged, first one that seeks the shortest path in the tangent bundle~\cite{Ben-Yosef_Ben-Shahar_2012_PAMI} and later the tangent bundle Elastica~\cite{Ben-Shahar_Ben-Yosef_2015_PAMI}, are very simple to describe (though not necessarily to solve) and provide visually pleasing curves. Like in previous models, no perceptual validation or evaluation against ground truth were yet performed.

While most of the prior art in curve completion sought the {\em perceptual} completed curve, it is also reasonable to seek the likely {\em physical} curve between inducers, a task that lends itself to  properties of real word image curves and natural image statistics. Indeed, some prior art did employ various statistics for the estimation of different aspects of natural images, showing that they follow properties such as scale invariance \cite{Ruderman_Bialek_1994_PRLetter}, or that the lengths of curve segments between two locally maximal curvature points follow a power low \cite{ren_malik_2002_ECCV}. 
Focusing on the co-occurrence of oriented edges, a strong preference of curve elements to co-linearity was shown by employing the statistics of edge pairs \cite{Geisler_etal_2001_VR}, and further high-order structure was shown by examining the statistics of edge triplets \cite{Lawlor_Zucker_2013_NIPS}.
In this paper we combine both schools, i.e., we explore natural image curve statistics for the task of long range curve reconstruction that best fits the physical reality (rather than, for example, the perceptual process or its outcomes).

Our investigation of the statistics of curves follows the framework of active shape models \cite{Cootes_etal_1995_CVIU}, employing the mean and variance of point correspondences and allowing a simpler investigation of shape variability relative to more complex methods for averaging curves \cite{sebastian_etal_2000_curve_atlas}.

\section{Global Curve Statistics - Analysis and Reconstruction}
\label{sec:model}

Given an input inducer configuration $\mathcal{C}=(\mathbf{I^1},\mathbf{I^2})$ for inducers $\mathbf{I^1}=(I^1_x,I^1_y,I^1_\theta)$ and $\mathbf{I^2}=(I^2_x,I^2_y,I^2_\theta)$ at locations $\mathbf{I^1_{xy}}=(I^1_x,I^1_y)$, $\mathbf{I^2_{xy}}=(I^2_x,I^2_y)$ and orientations $I^1_\theta$, $I^2_\theta$, we seek to generate a curve represented as a set of $n$ points $\alpha(\mathcal{C})=\{ \mathbf{x_1},...,\mathbf{x_n}  \}$ that closely matches the way physical visual (i.e., image) curves behave between such two inducers. That latter behavior will be measured from a large collection of observable image curves. 

For this data driven reconstruction we suggest to employ the distribution of natural curves for a given configuration:
\begin{align}
	\label{eq:main}
	P(\mathbf{x_1},...,\mathbf{x_n} | \mathbf{I^1},\mathbf{I^2})   \,\,  .
\end{align}
Previous works that employed the statistics of curves sought the most likely completion, which they were able to calculate by making assumptions that enable to express this distribution with smaller functions that capture local statistics of shape. Here, we suggest a global approach of estimating the mean curve instead, and as observed in Fig.~\ref{fig:summary}d, it may also provide a good estimation of the most likely curve.

In the following we describe the data collection and completion process, as well as its practical difficulties to generate proper reconstructions when the inducers are too far, among other conditions. To make the process more generally applicable, we show that the mean curve possesses certain properties that facilitate more stable and visually appealing completion even under those challenging conditions. 

\subsection{Collection and Representation of the Prior}
\label{sec:collection}

As exemplified in Fig.~\ref{fig:summary}, the basic reconstruction of a curve between a given inducer configuration $\mathcal{C}$ follows several steps. Note that most of the computations can and are done just once as preprocessing and need not repeat themselves for each reconstruction query, which essentially can be answered by a lookup operation.

As a first step, we collect a set of natural image curve {\em fragments} with configurations that are similar to $\mathcal{C}$. To that end, we employ the existing Curve Fragment Ground-Truth Dataset (CFGD) \cite{guo_kimia_2012_CVPRW} that contains ground-truth annotations of perceived image curves collected from three different annotators on highly varied scenes from the Berkeley Segmentation Dataset (BSDS) \cite{Martin_etal_2001_ICCV}, leading to \texttildelow40K perceived curves represented as a set of ordered points. Via proper processing we consider any sub-curve of each CFGG curve as a fragment in its own right, an operation implemented by choosing any possible pair of points along a given CFGD curve as starting and ending points of a fragment. This process provides a total of \texttildelow19M fragments.

Next, to make these fragments useful for our statistical analysis, we represent them by the {\em relative configuration} of their inducers:

\begin{small}
	\begin{align}
		\mathcal{C}_r=\mathbf{p}=(p_x,p_y,p_\theta)=(I^1_x-I^2_x,I^1_y-I^2_y,I^1_\theta-I^2_\theta)
	\end{align} 
\end{small}
where $\mathbf{p_{xy}}=(p_x,p_y)$ and $p_\theta$ are the location and orientation of the second inducer relative to the first one.
In particular, we translate and rotate the curve fragment with configuration $\mathcal{C}$ such that the location of its first inducer $\mathbf{I^1_{xy}}=(I^1_x,I^1_y)$ overlaps the origin of the coordinate system, and its orientation $I^1_\theta$ coincides with the $X$ axis. This operation effectively assumes invariance of shape to translations and rotations and allows a representation of the transformed fragments by the second inducer only.
Assuming also invariance to reflection, we finally reflect all fragments with positive relative elevation (i.e., $p_y>0$) so all curves end in the bottom half of the coordinate system. Finally, for a quick lookup of fragments later on, they are stored according to their relative configuration $\mathbf{p}$. All this pre-processing is exemplified in Fig.~\ref{fig:summary}a,b.

To extract the probability distribution of natural curves that match $\mathcal{C}$, we now collect all curve fragments that fit the corresponding relative  inducer configuration, allowing some tolerance in position and orientation. Formally, we include all curve fragments with relative configuration $\widetilde{\mathcal{C}_r}=\mathbf{q}$, such that \mbox{$\frac{\norm{\mathbf{p_{xy}}-\mathbf{q_{xy}}}}{\norm{\mathbf{p_{xy}}}}<t_1$} and $d_\pi(p_\theta,q_\theta)<t_2$, where $d_\pi$ is the angular distance between two angles.
That is,  we include fragments if their inducer configurations deviate from $\mathbf{p_{xy}}$ only marginally in normalized distance {\em and} in relative orientation, where the margin is determined by two predefined small thresholds $t_1$ and $t_2$. 
Note that since the collected fragments have similar but not identical configurations, they are further finely transformed to allow meaningful pooling in the reconstruction step. In practice, they are just slightly scaled so their end point $\mathbf{q_{xy}}$ exactly matches $\mathbf{p_{xy}}$. In the interest of space, the few trivial technical details of this step are listed in the supplementary materials\footnote{Supp. material is provided in \url{http://icvl.cs.bgu.ac.il}}.  Note that small discrepancies in orientations $q_\theta$ and $p_\theta$ may still persist but do not affect the reconstruction process.

\subsection{Basic Curve Reconstruction}

As the reconstruction of missing curve fragments is based on the ``mean curve'', the latter must be represented in a convenient way. For configuration $\mathcal{C}$ with relative configuration $\mathcal{C}_r$,  we represent transformed fragments in the collected data as a set of $n$ points $\alpha_i(\mathcal{C}_r)=\{ \mathbf{x_1},...,\mathbf{x_n}  \}$ sampled from the original points such that the arc length between each two adjacent points in $\alpha_i(\mathcal{C}_r)$ is equal (blue points in Fig.~\ref{fig:summary}b). The mean curve $\alpha(\mathcal{C}_r)$ is estimated as:
\begin{align}
	\alpha(\mathcal{C}_r) = \frac{  1  }{m(\mathcal{C}_r)} \sum_{i=1}^{m(\mathcal{C}_r)}   \alpha_i(\mathcal{C}_r)    \,\,   ,
\end{align}
where $m$ is the number of available fragments $\alpha_i$ for relative configuration $\mathcal{C}_r$ (Fig.~\ref{fig:summary}c). Finally, $\alpha(\mathcal{C}_r)$ is translated and rotated back to configuration $\mathcal{C}$, providing $\alpha(\mathcal{C})$. The reconstructed curve as a whole is obtained by interpolating these points, either linearly (as in Fig.~\ref{fig:summary}c) or using more elaborate methods.

As exemplified in Fig.~\ref{fig:summary}, this process often provides intuitive, visually pleasing, and veridical  reconstructions. Such results come as no surprise once we examine the nearly normal distribution of the curve sample points, for example the one for the central curve point depicted in Fig.~\ref{fig:summary}d.  Note that this happens despite the observation that the center point, being ``furthest'' away from both inducers along the arc length, is the one where curves are most likely to show the greatest variation.

\begin{figure}[h]
	\begin{center}
		\begin{tabular}{cc}
			\includegraphics[width=0.49\linewidth]{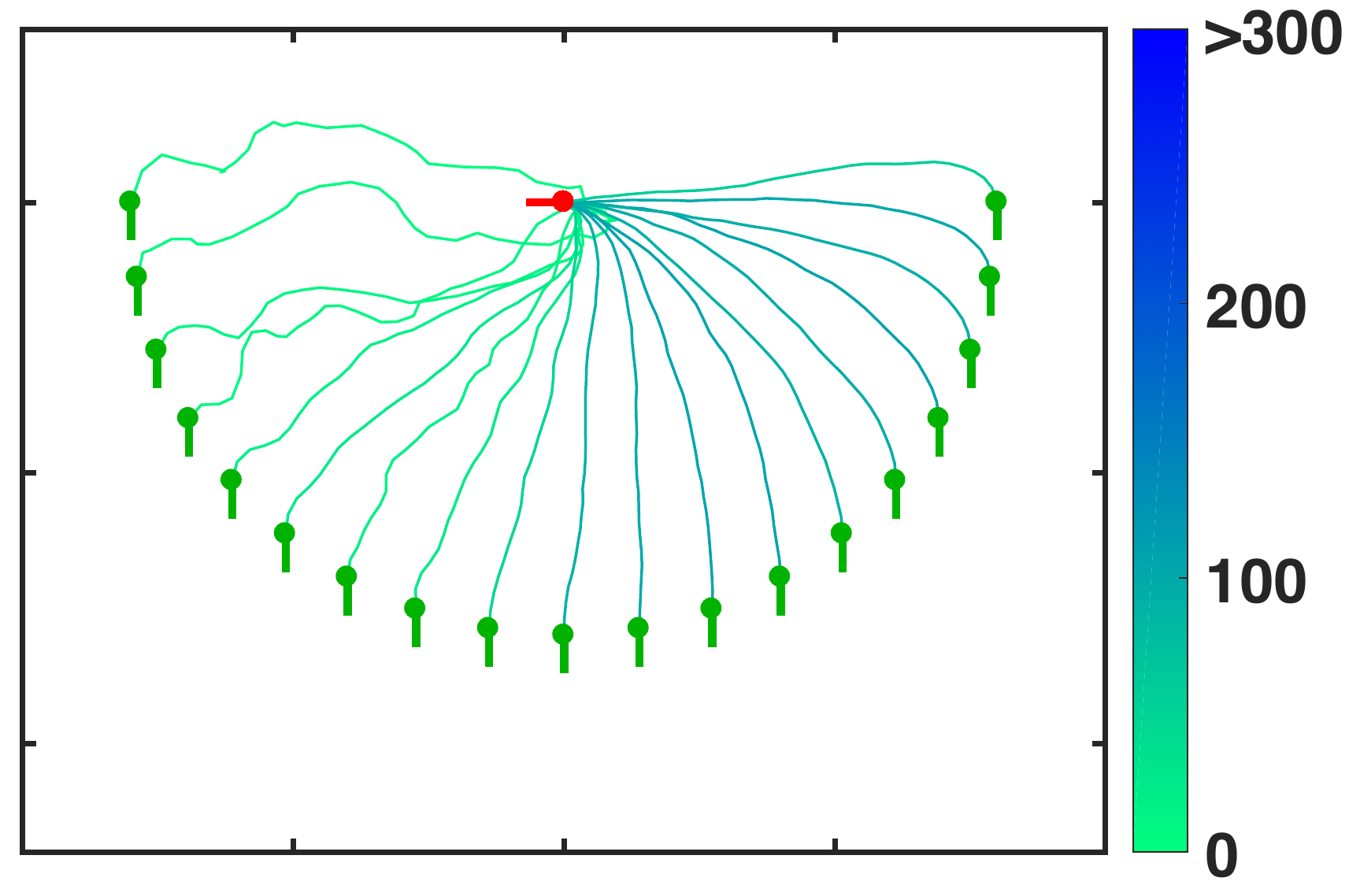} &
			\includegraphics[width=0.49\linewidth]{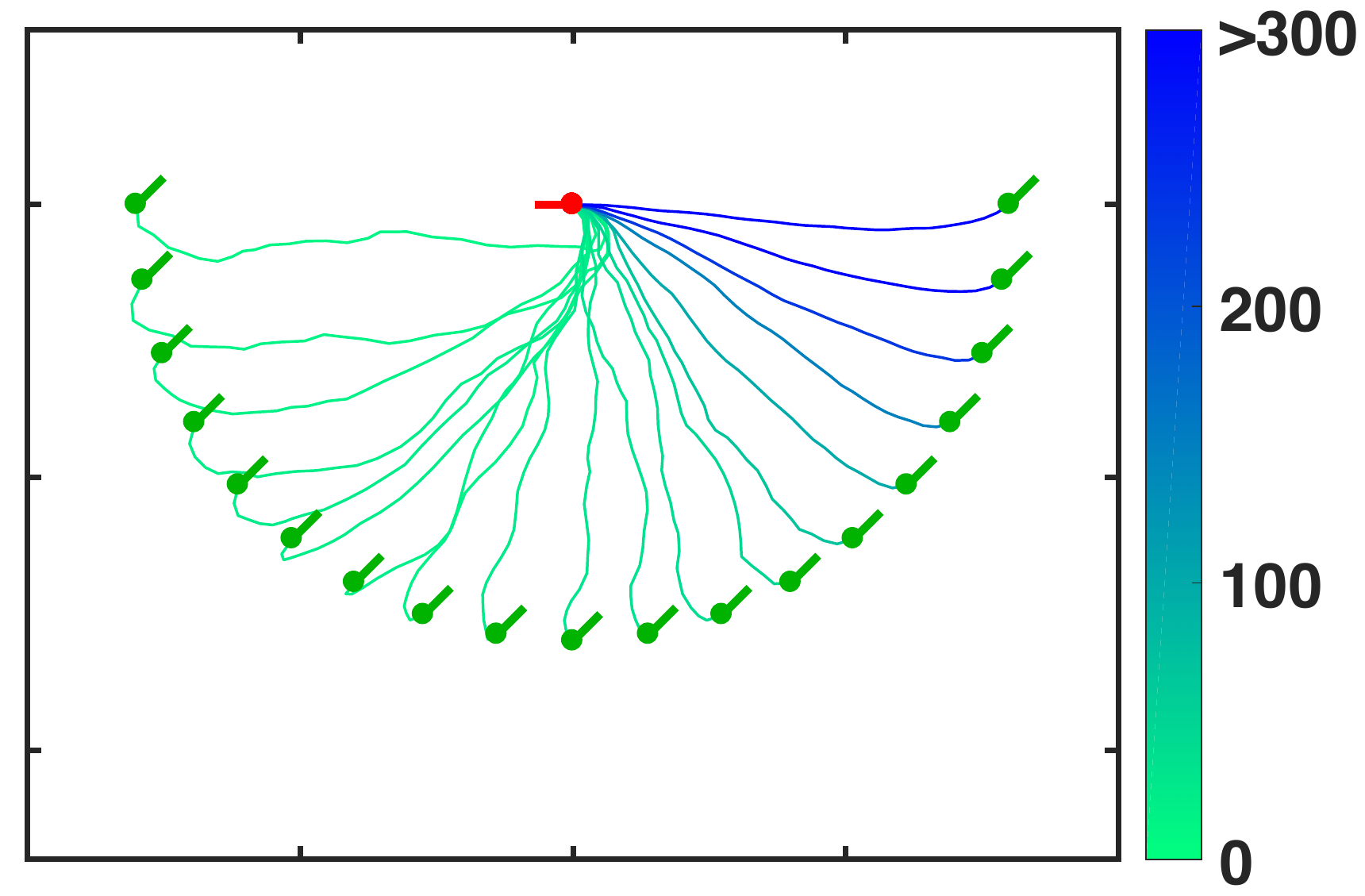} \\
			(a)&(b)\\
		\end{tabular}
	\end{center}
	\caption{Basic reconstructions of different inducer configurations based on the mean curve. All configurations share one inducer (red) but have different second inducers (green), in this case at the same distance (80 pixels).  Curve colors represent the number of corresponding fragments that were found in the dataset and thus used during reconstruction. Note how larger sample sets allow more intuitive and visually pleasing reconstructions. 
	}
	\label{fig:results_mean_curve}
\end{figure}

That said,  and as Fig.~\ref{fig:results_mean_curve} shows, the quality of results highly depends on the number of ground truth curve fragments that are observed in the prior (denoted as darker blue curves), or more formally, on the power of the statistics for any given inducer configuration. Empirically, reconstructions appear noisy when sample size is small, a situation more common for example when inducers are oriented away from each other, especially for larger gaps (i.e. long curves). Clearly, one way to address this difficulty is collecting more data and annotating many more curves, two operations that require much human labor. However, one can do much better even with existing data (in our case, the CFGD collection~\cite{guo_kimia_2012_CVPRW}) and indeed, to alleviate the problem we now examine and exploit certain properties of the mean curve that allow ``sharing'' fragments between configurations, boosting the number of examples and the power of the statistics, and thus greatly improve the results in the ``problematic'' cases also.

\subsection{Boosting via Invariance to Scale}
\label{sec:scale_inv}

If curve fragments (or curves in general) were scale invariant, curves (either physical or reconstructed) would scale similarly to their inducer configuration. Hence, if we could obtain evidence to that effect, the distribution of curve fragments at one scale could help determine the reconstructions at other scales also and the scarcity of samples for certain configurations, especially for longer curves, would be alleviated. Clearly, with such a capacity, one would even be able to generate reconstructions  of arbitrarily long curves at arbitrary scales even if such curves were {\em never} observed in the prior. 
In this section we show that the mean curve is scale invariant in most cases, facilitating data boosting and better reconstructions in general. We note that these findings are congruent with earlier natural image statistics \cite{Ruderman_Bialek_1994_PRLetter,Geisler_etal_2001_VR}, though here we show them for the shape of whole curves. For the interest of space we focus the presentation on the curve center points.

Loosely speaking, the reconstructed curve for some inducer configuration is invariant to scale if it is similar to the reconstruction at any other scale after proper scaling of the configuration.  Since scaling is analogous to changing viewing distance, and since such an operation does not change the observed orientation, scaling essentially affects only the distance between inducers, and the ``scale'' $s$ of a curve fragment is determined by that distance alone.  

\begin{figure*}[h!]
	\begin{center}
		\begin{tabular}{ccc}
			{\small Horizontal inducers} & {\small STD of $\mu_s$}  & {\small Average of $\sigma_s$}  \\
			\raisebox{1.5mm}[0pt][0pt]{\includegraphics[width=5cm]{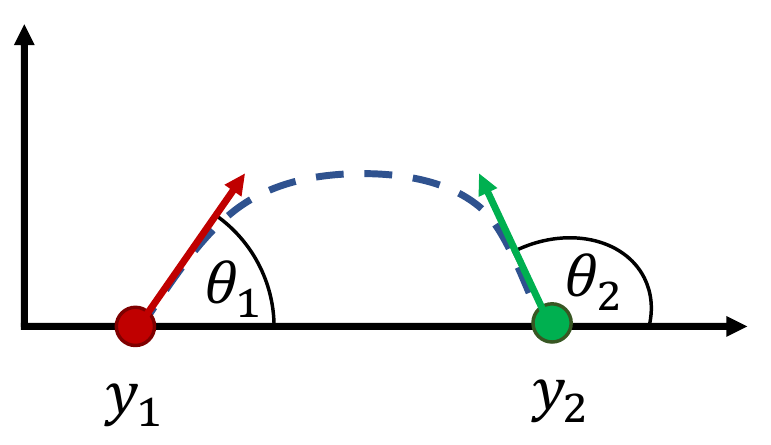}} &
			\includegraphics[width=5.5cm]{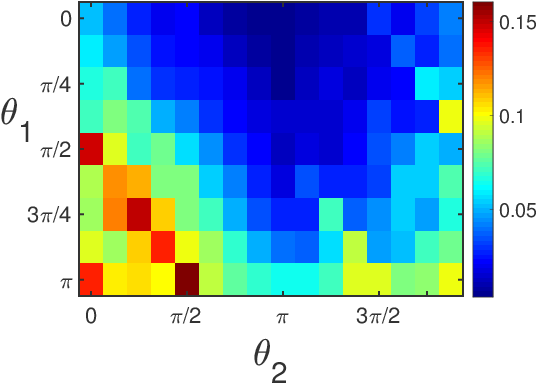} &
			\includegraphics[width=5.5cm]{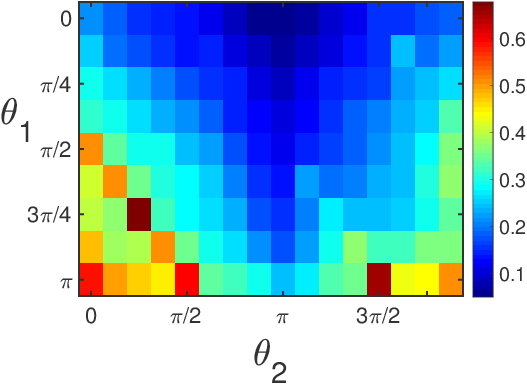} \\
			(a)&(b)&(c)
		\end{tabular}
		\caption{
			Empirical analysis of scale invariance for a subset of inducer configurations. 
			{\bf (a)} Considered are all pairs of horizontal inducers with two arbitrary orientations.
			{\bf (b)} A map of the STD of $\mu_s$ of the curve center point across different scales 
			shows very stable behavior when the inducers are ``facing" each other with gradually decaying stability as the inducers ``face" away from each other.
			{\bf (c)} The average of $\sigma_s$ across scales similarly shows very stable behavior when 
			the inducers roughly ``face" each other, gradually turning unstable as they ``face" away from each other. 
			The two maps in conjunction imply scale invariance for facing inducers, and no conclusive insights otherwise.
		}\vspace{-15pt}
		\label{fig:scaleInv}
	\end{center}
\end{figure*}

With this in mind, to examine the scale invariance hypothesis we explore the effect of scale on the reconstruction at a base relative configuration $\mathcal{C}_r=\mathbf{p}$ with nominal scale~1 (i.e., with $\norm{\mathbf{p_{xy}}}=1$).
At each scale $s$, fragments for relative configuration $\widetilde{\mathcal{C}_r}=(s\mathbf{p_{xy}},p_\theta)$ are collected and transformed as described in Sec. \ref{sec:collection}, and then brought to a common scale by transforming them once more such that the distance between their inducers is 1. Focusing on the center points of the transformed fragments from the original scale $s$, let $\mu_s$ and $\sigma_s$ be their mean and standard deviation. We summarize the information across scales as the standard deviation of the means $\mu_s$ (see Fig~\ref{fig:scaleInv}b ) and the expected value of the standard deviations $\sigma_s$ (Fig~\ref{fig:scaleInv}c). 

Since the STD of $\mu_s$ is the mean difference between the reconstructed central curve points across scales, one would expect it to be smaller the more invariant the curves to scale. In Fig.~\ref{fig:scaleInv} we examine this difference in the reference scale of 1. As shown, for ``normal'' inducer configurations the STD of $\mu_s$ is very small compared to the scale, suggesting that the mean curve is indeed invariant to scale in these cases. For other, more ``abnormal'' configurations we see greater STD, suggesting that the calculated mean curve is different across scales. Note that this in itself does not necessarily indicate that the mean curve is scale {\em variant} in such difficult configurations, for the average $\sigma_s$ in these cases is large as well (Fig.~\ref{fig:scaleInv}c), requiring more samples to guarantee an accurate estimation of the mean according to Chebyshev's inequality applied for the empirical mean \cite{mohri_etal_2012_foundationsOfML}.

Following this analysis, we now leverage the invariance of the mean curve to scaling (assuming scale invariance also when it could not be asserted) by extracting the distribution of natural curves that match $\mathcal{C}_r=\mathbf{p}$ from fragments $\widetilde{\mathcal{C}_r}=\mathbf{q}$ that are close to any of its scaled versions. Formally, we  include all curve fragments with relative configuration $\widetilde{\mathcal{C}_r}=\mathbf{q}$, such that the angular distance between vectors $\mathbf{p_{xy}}$,$\mathbf{q_{xy}}$ and orientation distance $d_\pi(p_\theta,q_\theta)$ are smaller than predefined thresholds $t_1$,$t_2$. To quickly collect such curves, fragments are stored ahead of time according to the angle to $\mathbf{q_{xy}}$ from the $X$ axis, and their orientation $q_\theta$. 

The process of utilizing curves at all scales greatly improves the results and now provides intuitive, visually pleasing, and more veridical reconstructions (see Fig.~\ref{fig:results}b)  for many more inducer configurations (Fig.~\ref{fig:results_mean_curve}). Another favorable property is the ability to generate curves regardless of the distance between given inducers. That said, it can be seen that the number of examples, for some ``difficult'' configurations remains small and thus the reconstruction in such cases is still noisy, a problem that would possibly be alleviated with the inclusion of additional samples. To this end, we now turn to explore yet another possible property of curves known as extensibility.

\subsection{Boosting via Midway Extensibility}
\label{sec:ext}

As discussed above, scale invariance alone does not suffice to increase the number of samples and qualify as a robust prior in all cases. Typically, the cases that remain problematic have their inducers face away from each other and so the mean curve bends rapidly near the inducers and keeps a straighter shape half way from them (Fig.~\ref{fig:results}b). Due to this behavior, another inducer placed in the center of such curve (and oriented along its tangent) would represent much smaller orientation relative to either of the inducers, effectively better ``facing'' both of them and thus forming less difficult configurations with either. As observed in Figs.~\ref{fig:results_mean_curve},\ref{fig:results}b, such simpler configuration tend to have many more dataset samples and smoother reconstructions. Therefore, generating a curve by reconstructing two sub-curves from the original inducers to the curve's center is likely to provide a more robust, visually pleasing and veridical reconstruction.

We refer to this scheme as {\em midway extensibility}, a special case of general extensibility, a property sought after in axiomatic approaches (\eg, \cite{Ullman_1976_BC}, cf.~\cite{Ben-Yosef_Ben-Shahar_2012_PAMI,Ben-Shahar_Ben-Yosef_2015_PAMI}). A reconstruction approach is considered extensible if for any pair of oriented inducers $\mathbf{I^3}$,$\mathbf{I^4}$ extracted from a reconstructed curve $\alpha$ between inducers  $\mathbf{I^1}$,$\mathbf{I^2}$, it generates a curve $\beta$ that is identical to $\alpha$ between $\mathbf{I^3}$,$\mathbf{I^4}$. 
Here, we explore the extensibility of the mean curve by its center point, and define a reconstruction scheme as \emph{midway} extensible if for any generated curve $\alpha$ between inducers $\mathbf{I^1}$,$\mathbf{I^2}$, $\alpha$ is identical to the curve $\beta=\beta_1\oplus\beta_2$, calculated as the concatenation of curves $\beta_1$ and $\beta_2$ that are generated between inducers $\mathbf{I^1}$,$\mathbf{I^3}$ and $\mathbf{I^3}$,$\mathbf{I^2}$, where $\mathbf{I^3}$ is the inducer extracted from the center of $\alpha$.

To investigate the property of midway extensibility in the context of our mean curve we employ the scale invariant procedure described in Sec.\ref{sec:scale_inv} to generate a curve $\beta$ (Fig.~\ref{fig:results}c) for each curve $\alpha$ in Fig.~\ref{fig:results}b by generating two scale invariant curves $\beta_1$, $\beta_2$ from the original inducers to the center of $\alpha$. While this was done for {\em all} inducer configurations, to determine whether the mean curve exhibits midway extensibility we  compare the corresponding reconstructions $\alpha$ and $\beta$ only in cases where $\alpha$ was generated with many observations, as these are the only cases where the reconstruction is likely to be precise in the first place. A careful inspection of their shapes reveals that while some differences between such corresponding curves can be observed, they are typically insignificant relative to the scale of the curve. To measure the deviation from extensibility we calculate the Fr\'{e}chet distance \cite{eiter_and_mannila_1994_frechet} relative to the distance between inducers for corresponding curves in Fig.~\ref{fig:results} and in the complementary figure in the supplementary material, covering the range of relative configurations. Considering only curves for which more than 400 fragments were observed, the maximal distance between such curves is $4.9\%$ of the distance between inducers, and the mean is $2.5\%$. Thus, curves generated by employing midway extensibility do not vary much from their counterparts that were accurately reconstructed without extensibility. 

Establishing such empirical  midway extensibility, we reiterate that its sought-after effect is for cases where relative configurations enjoy only few samples in the prior. Checking the reconstruction facilitated by the two sub curves now shows significant improvement  in such cases (compare the relevant cases in Fig.~\ref{fig:results}b,c).

\section{Experimental Results}
\label{sec:results}

\begin{figure*}[h]
	\begin{center}
		\begin{tabular}{ccc}
			{\small Euler Spiral} & {\small Scale invariant (SI) Mean}  & {\small SI Extensible Mean}  \\
			
			\raisebox{0.6mm}[0pt][0pt]{\includegraphics[width=0.262
				\linewidth]{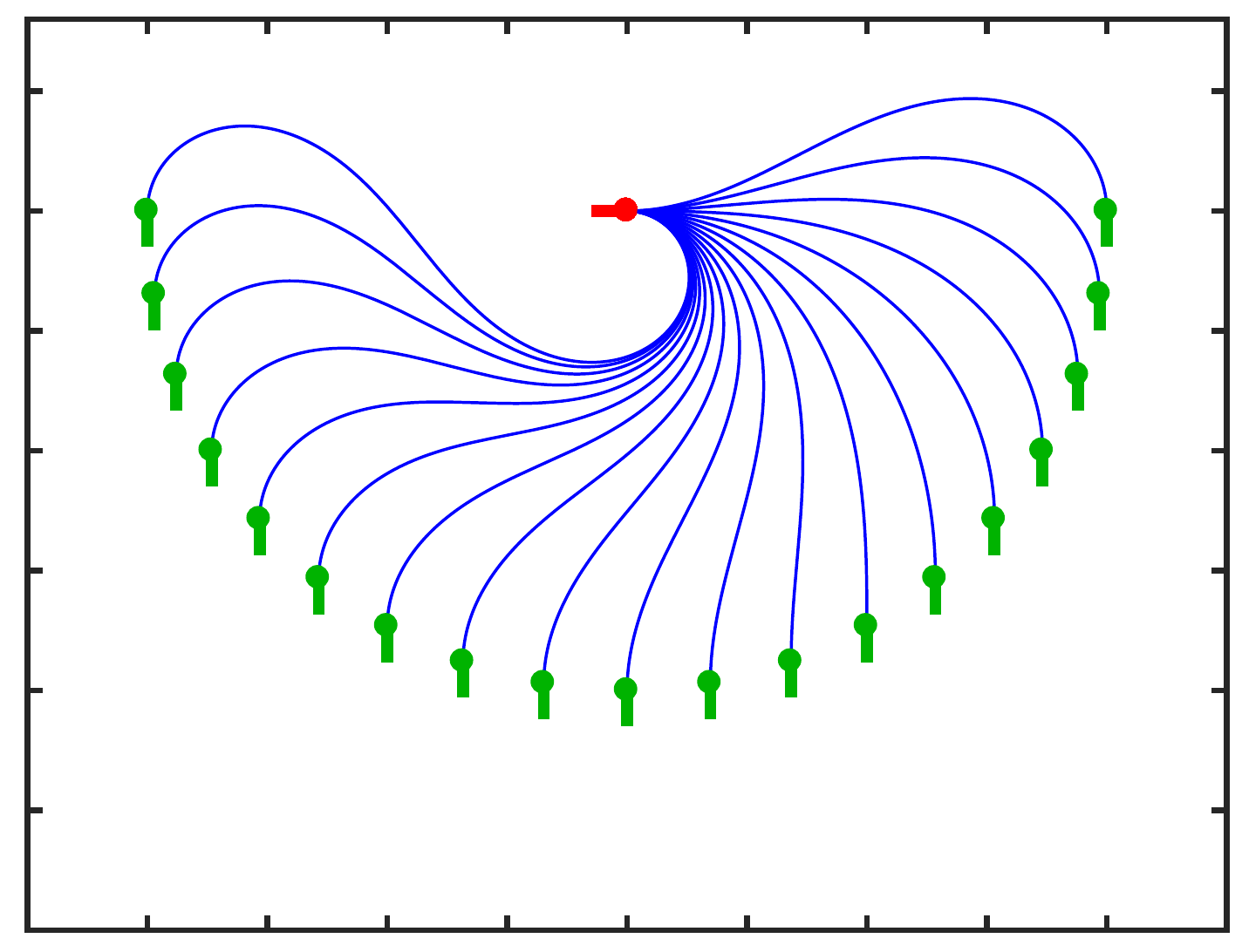}} &
			\includegraphics[width=0.32\linewidth]{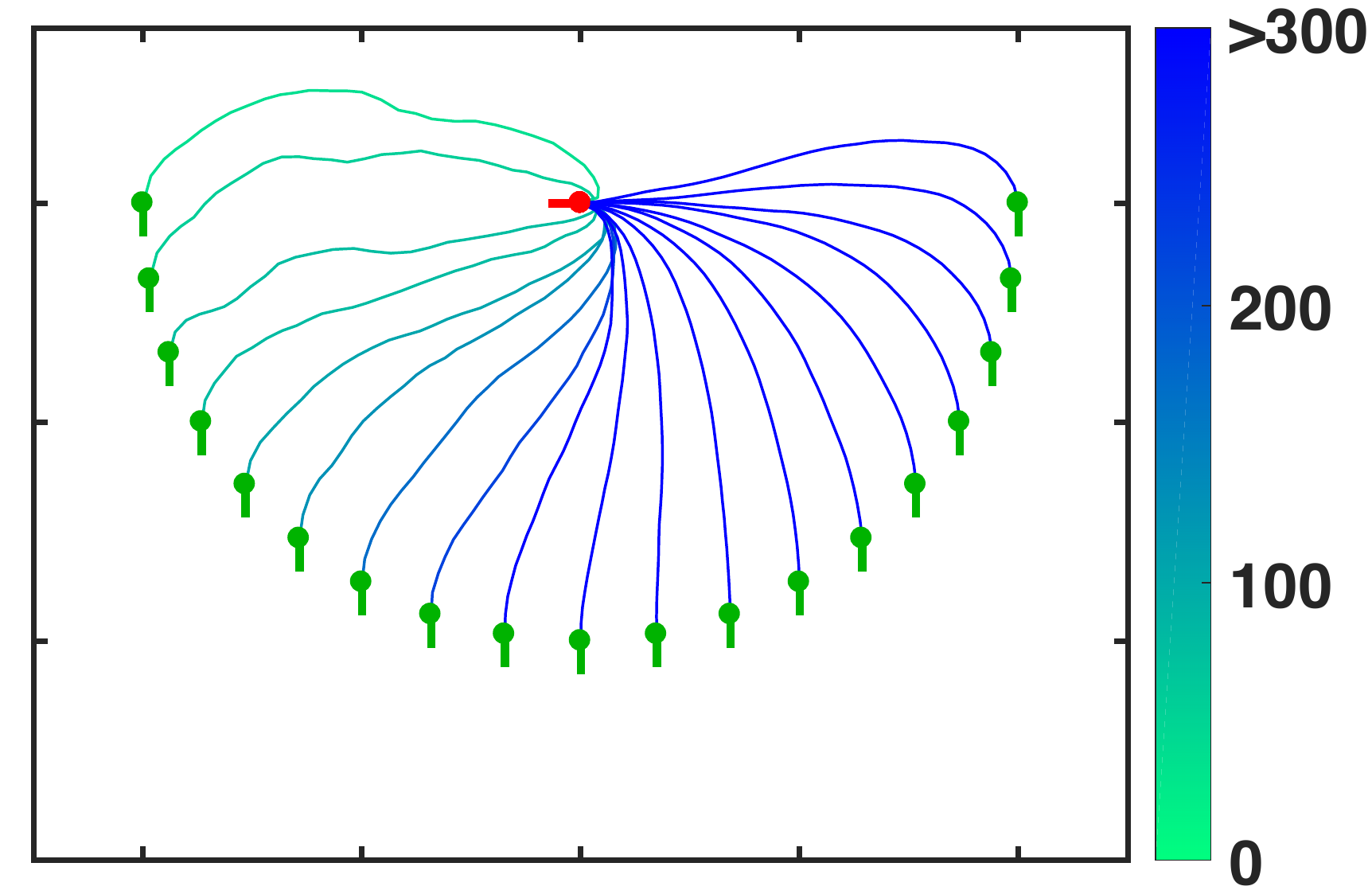} &
			\raisebox{-0.2mm}[0pt][0pt]{\includegraphics[width=0.32\linewidth]{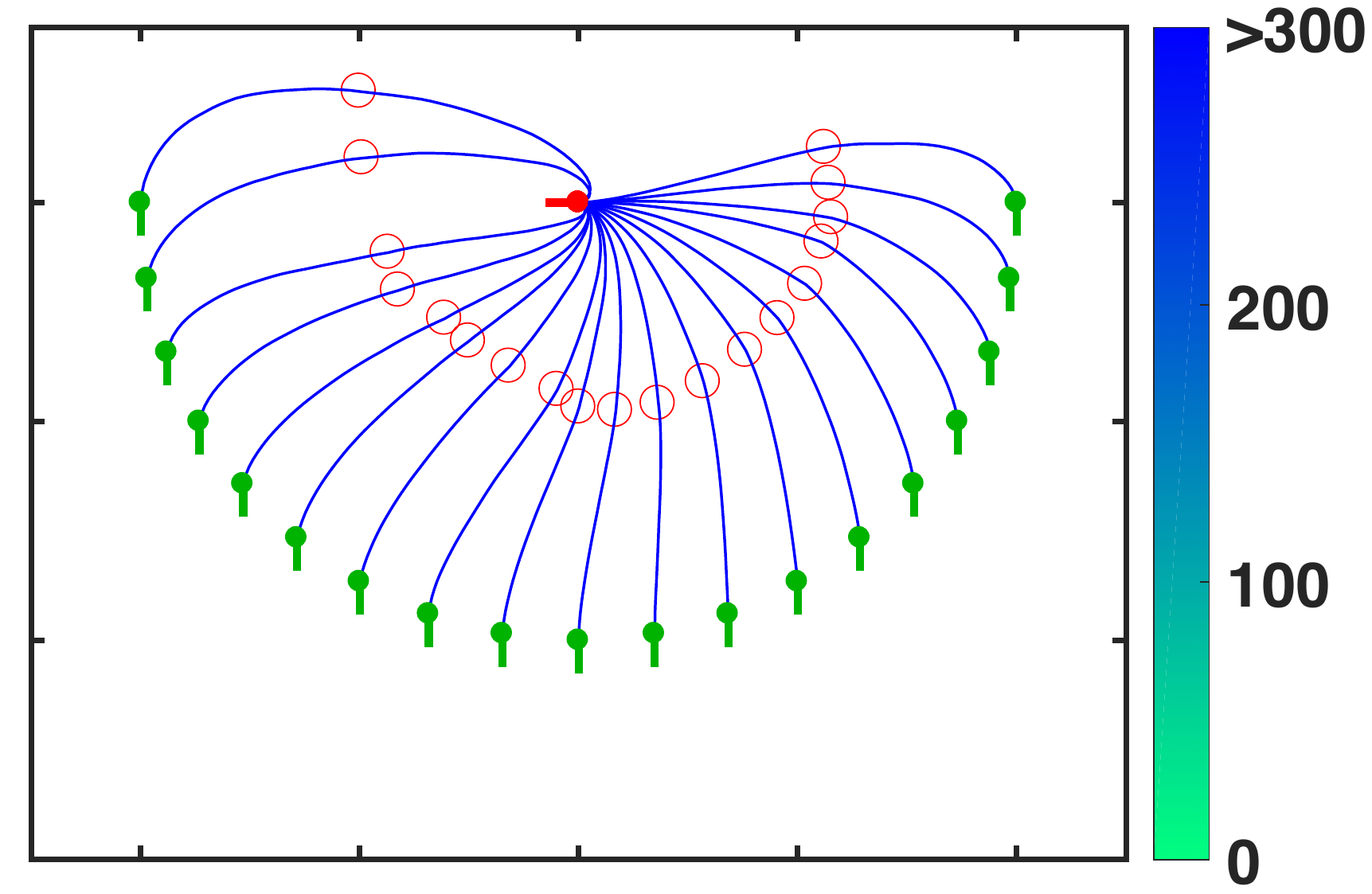}} \\
			
			\raisebox{0.29mm}[0pt][0pt]{\includegraphics[width=0.265\linewidth]{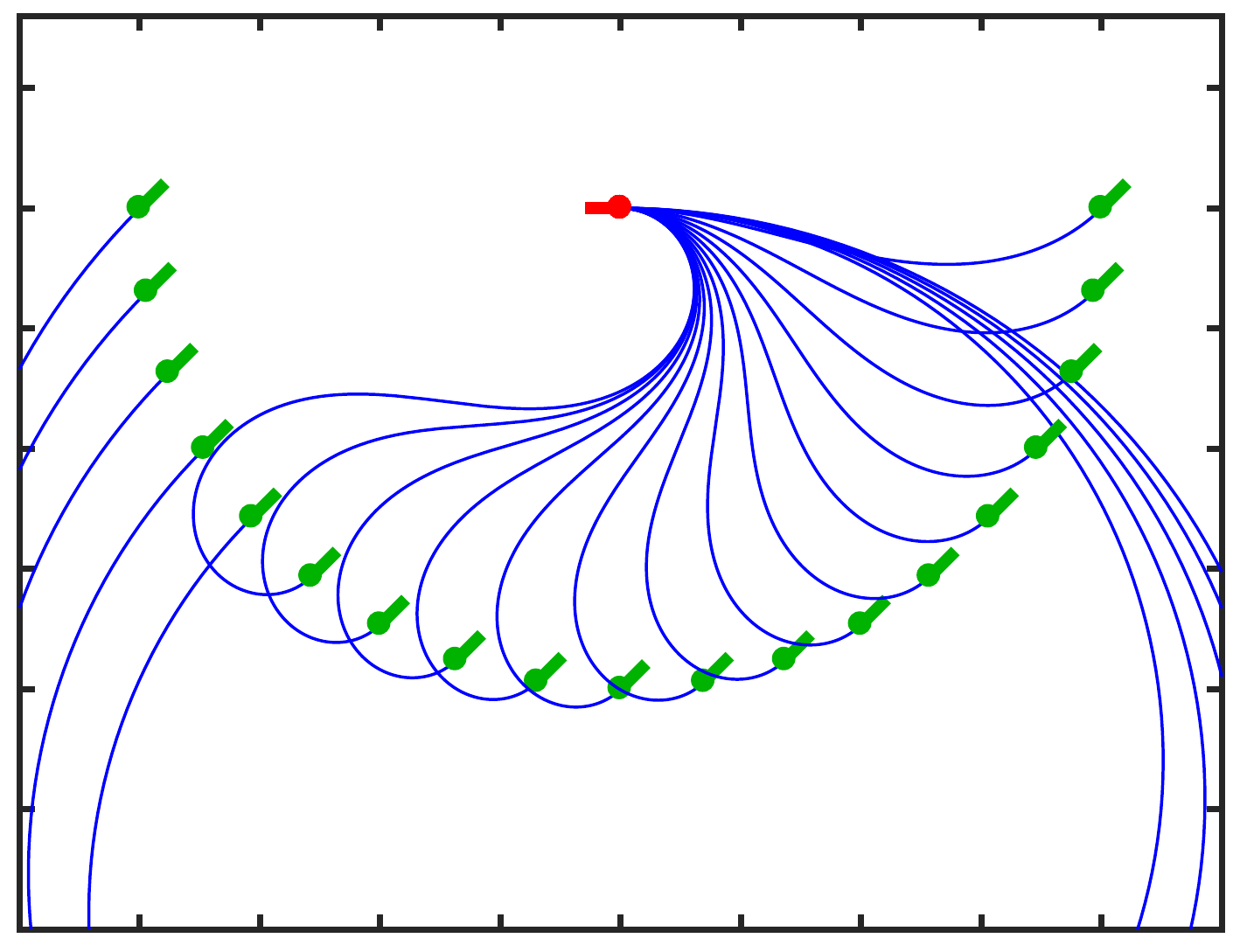}} &
			\includegraphics[width=0.31\linewidth]{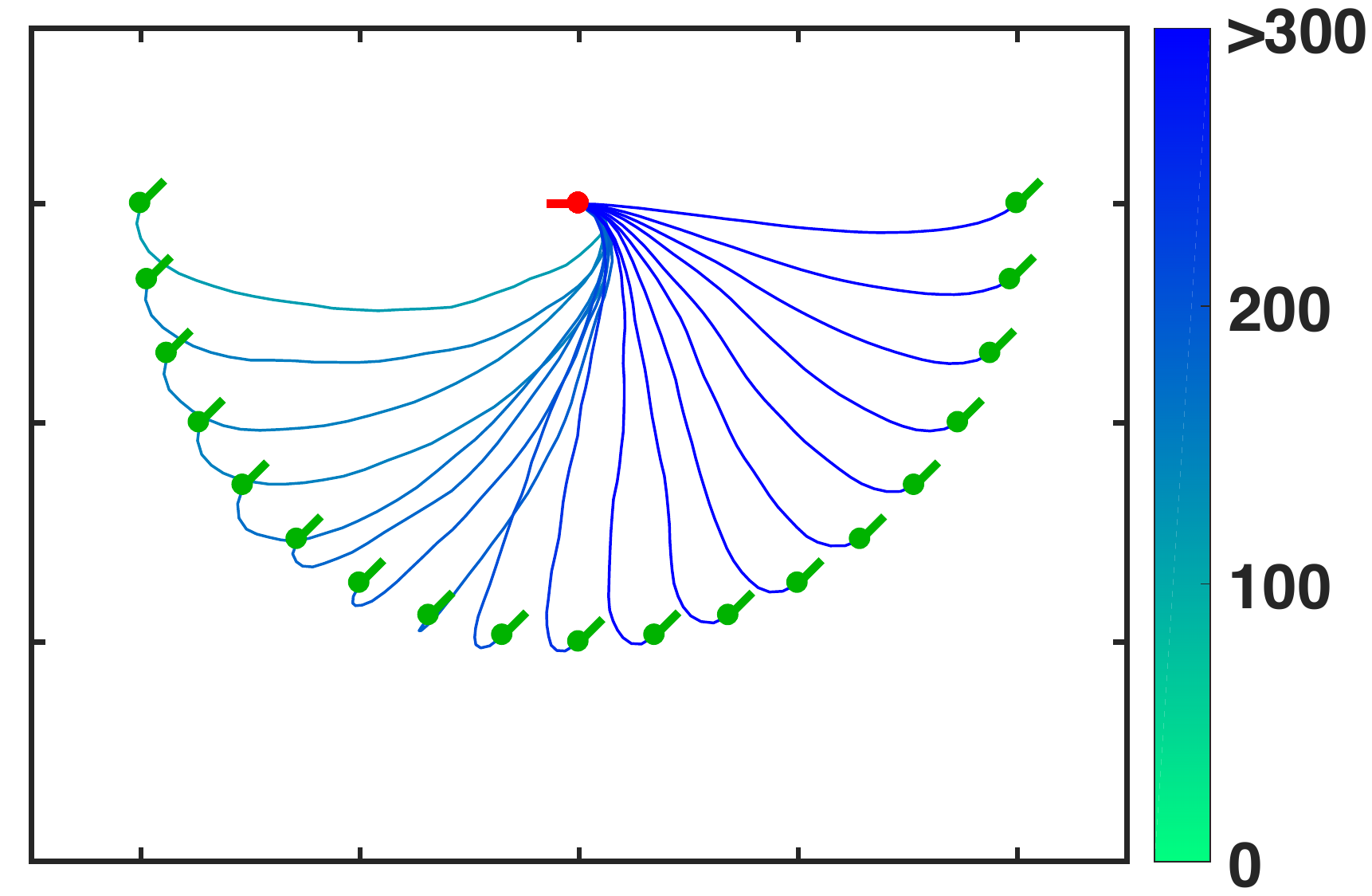} &
			\includegraphics[width=0.31\linewidth]{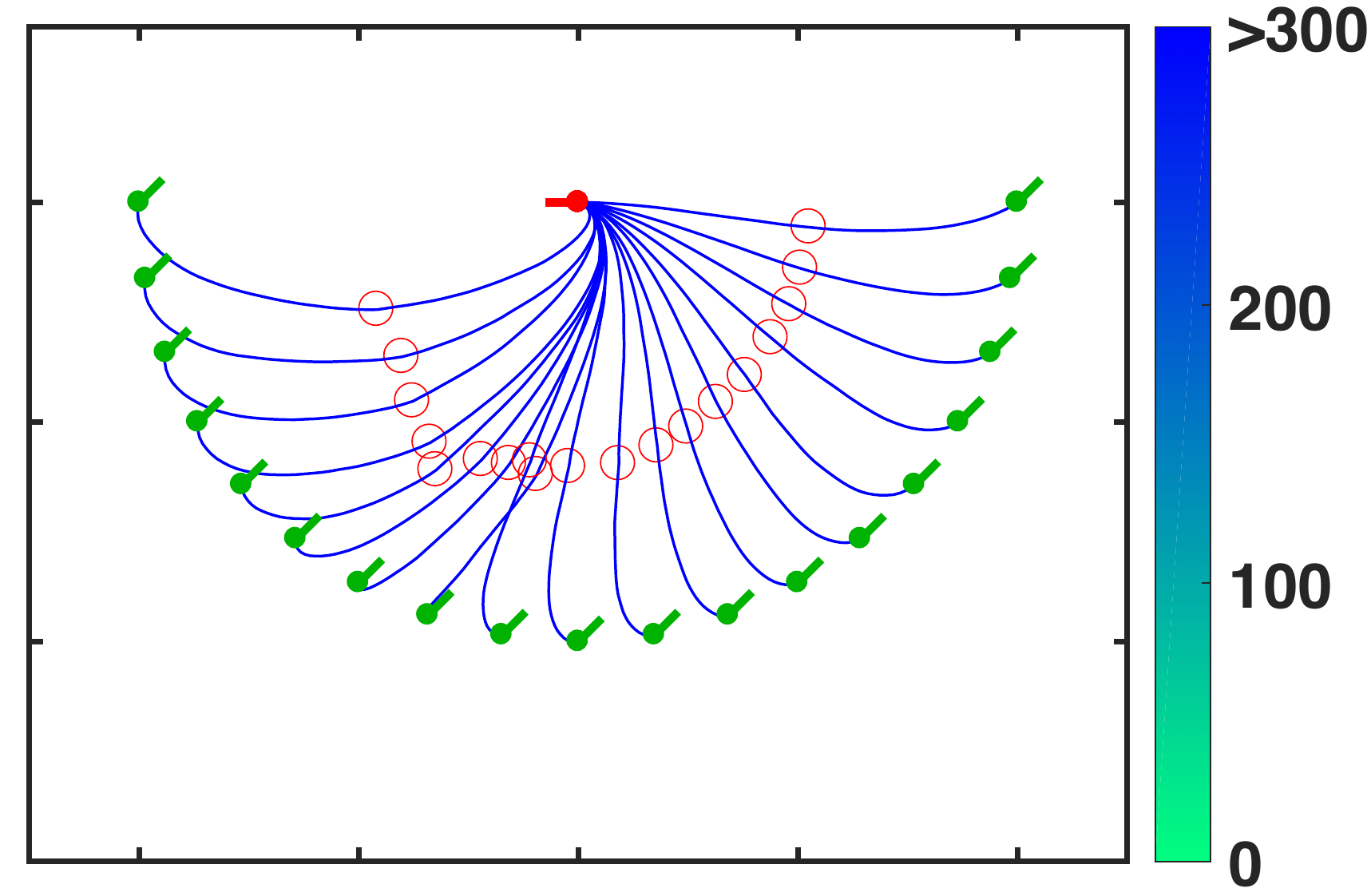} \\
			
			(a)  & (b)   &(c)  \\
		\end{tabular}
	\end{center}
	\caption{Reconstruction results based on (a) Euler spiral, (b) scale invariant mean curve, and (c) extensible and scale invariant mean curve, where darker curve colors represent a larger number of samples from which the curve was calculated. Red circles in (c) mark the points in which two reconstructions are connected to a single curve.  As exemplified, the final mean curve is not only data driven and thus more veridical, it is also more compact and it exhibits much smoothness even though no regularization or smoothing was performed.
	}
	\label{fig:results}
\end{figure*}

The suggested model calculates the mean curve using fragments from very similar configurations and across scales, where midway extensibility is employed if less than 400 fragments were collected. We compare our model with Libcornu's implementation of Euler Spiral \cite{xu_etal_2013_CVPR,walton_etal_2009_JCAM}, the latter being a prominent method for the completion of curves.

\begin{figure*}
	\begin{center}
		\begin{tabular}{ccc}
			\includegraphics[width=0.30\linewidth]{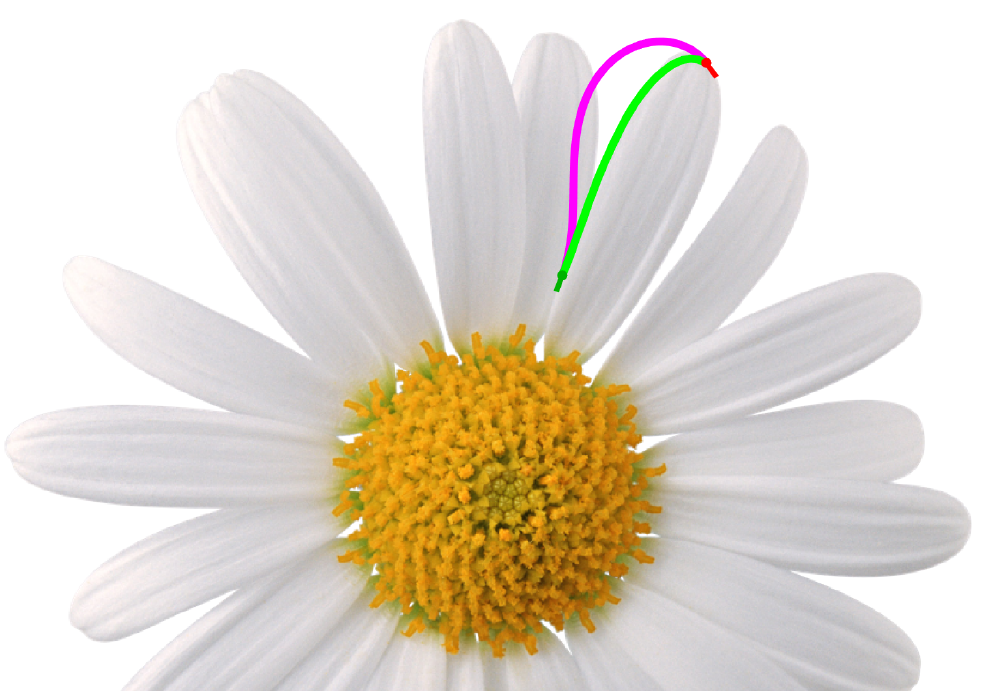} &
			\includegraphics[width=0.215\linewidth]{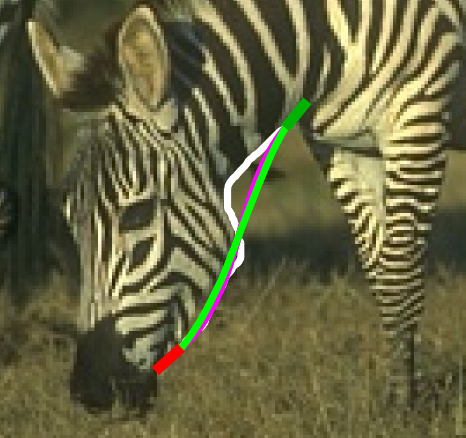} &
			\includegraphics[width=0.168\linewidth]{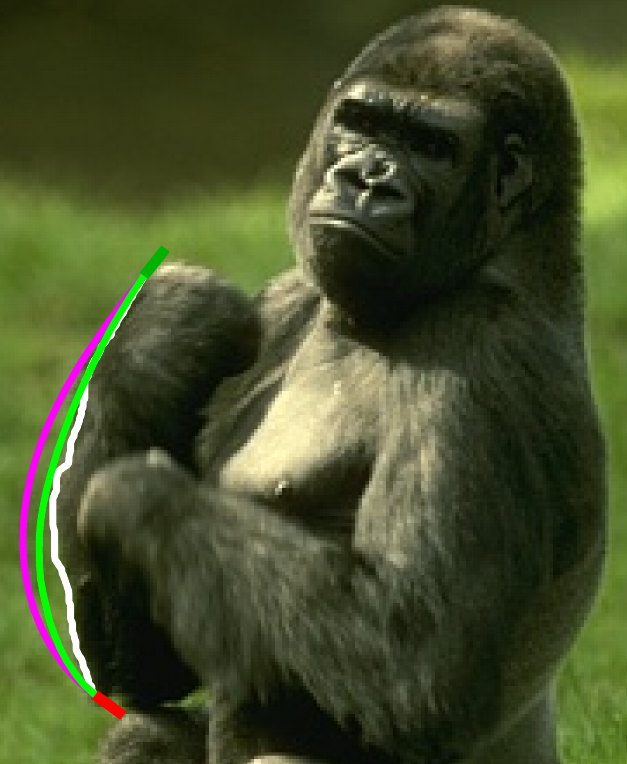} \\
			(a) & (b) & (c) \\
		\end{tabular}
	\end{center}
\vspace{-5pt}
	\caption{Example reconstructions over large gaps of possibly occluded curves based on our approach (green) and Euler spiral (magenta). Examples (b) and (c) are taken from the reconstruction benchmark, with the ground-truth curves shown in white. Image parts below the curves are assumed to be completely occluded and are not used for reconstruction.}
	\label{fig:reconstructions}
	\vspace{-7pt}
\end{figure*}

Reconstruction for all selected curve configurations can be seen in Fig.~\ref{fig:results} and in the supplementary material. Results are shown at a single scale and for the lower half visual field, a consequence of the scale and mirror reflection invariance properties of both models.  Comparing the results visually suggests that the Euler spiral, being an analytic regularized model, is generally smoother,  but it often provides reconstructions that appear unnatural and/or far from veridical (in a sense of corresponding to the actual physical curve as represented by the prior). When it comes to reconstruction of missing physical curve fragments our data driven approach thus provides an intrinsic advantage. A demonstration to that effect can be shown with the real objects in Fig.~\ref{fig:reconstructions}.

\subsection{Reconstruction Evaluation}

\begin{figure*}[t]
	\begin{center}
		\includegraphics[width=0.4\linewidth]{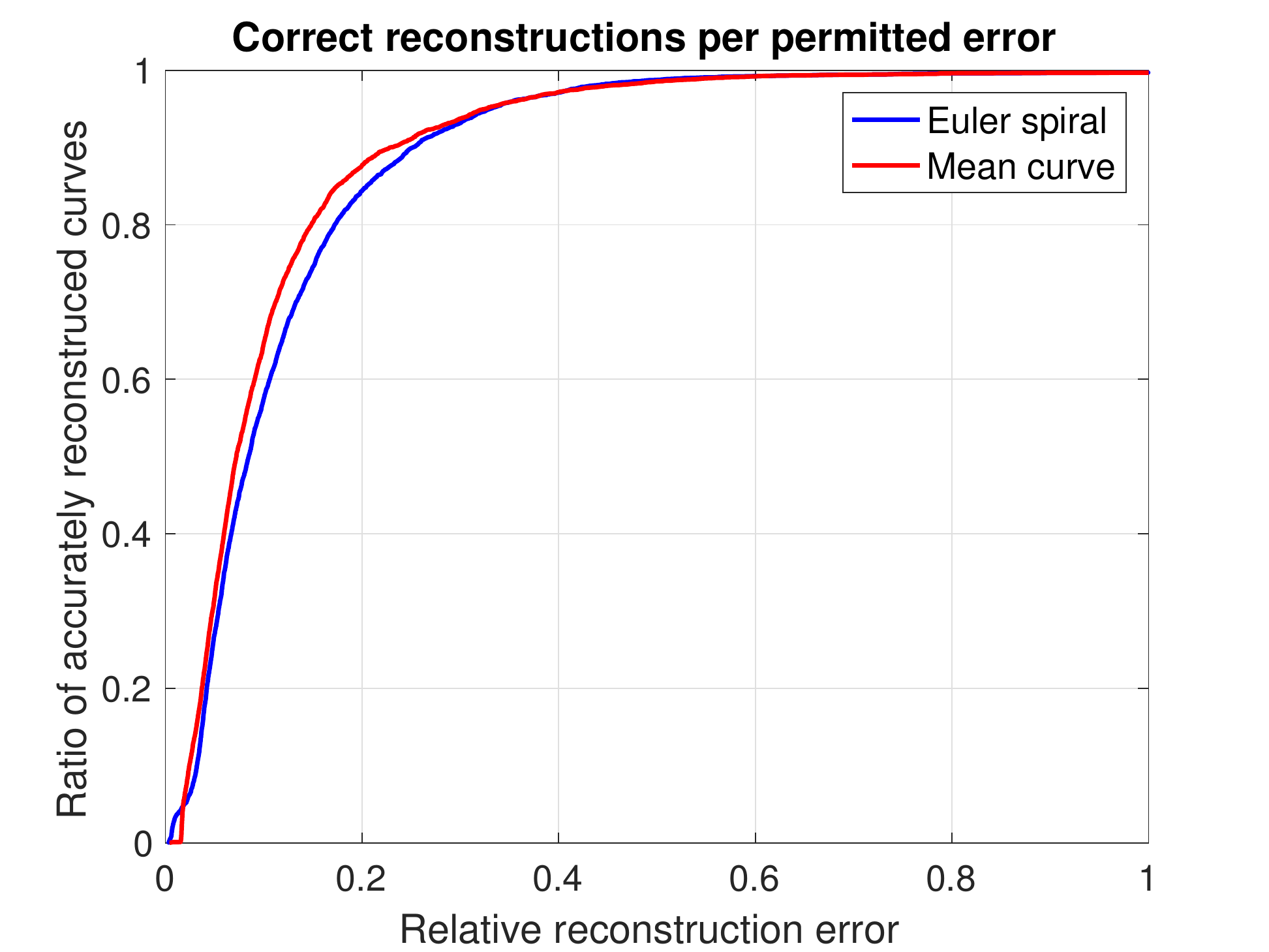}
		\includegraphics[width=0.4\linewidth]{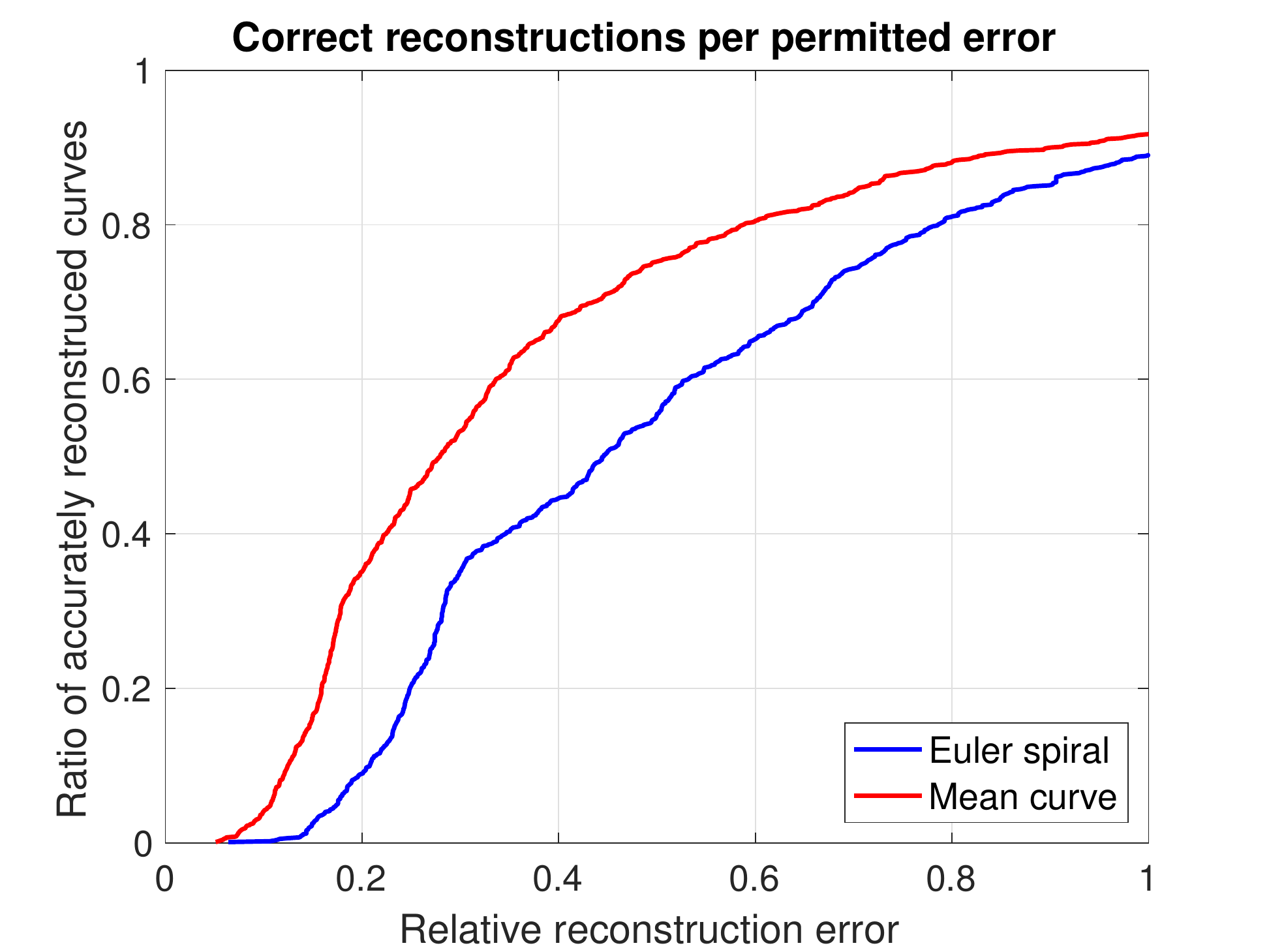}
	\end{center}
	\vspace{-5pt}
	\caption{
		Reconstruction evaluation showing the ratio of accurate reconstructions for each accuracy threshold defined according to the relative reconstruction error over the suggested benchmark (left) and for difficult inducer configurations (right). As shown, the graph of the mean curve dominates over the graph of Euler Spiral completions. 
	}
	\label{fig:results_reconst}
		\vspace{-7pt}
\end{figure*}

In addition to the qualitative results discussed above, we examined the accuracy of the mean curve as a reconstruction scheme for natural occluded curves. In our representation, the mean curve is, by construction, the one ``closest'' to all observable $n$-point natural planar curves (abstractly represented as vectors in $2n$ dimensional space). However, uniquely shaped curves do occur in nature and may be far from the mean. The extent to which such curves can be reconstructed, and the information or the prior needed to facilitate such reconstructions, remain to be studied. That said, we still seek to obtain a quantitative measure of a method's ability to properly reconstruct fragments of natural curves. Since the BSDS and CFGD benchmarks are not suitable for this task, we employ the data from CFGD and create a benchmark for the reconstruction of curves at different scales (which can be seen as differently sized gaps or holes in tasks such as image inpainting).
To do so, we first split the original dataset into training and test images, the latter comprises 10\% of the dataset. No curve fragments from test set images were used when computing the mean curve for reconstruction, while no curve from the train set was used for queries or as ground truth for performance evaluation.

More specifically, a set of 5000 curve fragments and their inducers were randomly selected from the test images such that fragments are uniformly distributed by their scale (as the distance between their inducers) to better represent a scenario such as the reconstruction of curves over randomly sized gaps.
For each ground-truth curve $\alpha^*$  from this test set, we then generated the mean curve  \mbox{$\alpha=\{y_1,...,y_n\}$} that matches its inducer configuration $(\mathbf{I^1},\mathbf{I^2})$  while using the fragments and the prior collected {\em only from train images}. We then evaluate the relative reconstruction error (RRE) of the generated curve $\alpha$ as the Fr\'{e}chet distance $d$ \cite{eiter_and_mannila_1994_frechet} between both curves relative to the distance between their inducers:
\begin{align}
	RRE=\frac{   d(\alpha^*,\alpha)   }{  \norm{\mathbf{I^1_{xy}}-\mathbf{I^2_{xy}}}    }  \,\,  .
\end{align}

Reconstruction results are shown in Fig. \ref{fig:results_reconst}, where each point in the $X$ axis represents an RRE threshold and its corresponding value in the $Y$ axis represents the ratio of accurately reconstructed curves (ARC) for which the error is less than that threshold. Overall results are summarized by the area under the curve (AUC), calculated as the average ARC at 101 RRE points from 0 to 1 (where the limit of 1 RRE was chosen arbitrarily). 
As can be seen, the mean curve provides more accurate reconstructions than Euler Spiral, with AUC scores of 0.887 and 0.876, respectively. The only cases where the Euler Spiral is marginally more accurate are those with extremely low threshold $RRE<0.017$ which, as we verified, comprises of straight lines only.

While one may argue that the differences in performance in Fig. \ref{fig:results_reconst} (left) are not big, it is also suspected that many of the details are washed by the frequent cases of ``convenient'' inducer configurations where both approaches yield similar results. Indeed, observing the different reconstructions such as in Fig. \ref{fig:results}, it can be seen that the mean curve and Euler Spiral differ more for the ``difficult'' configurations that are underrepresented in our dataset and in the prior. Therefore, we conducted a focused evaluation with an additional random set of 1000 curves, all with ``difficult'' configurations such that $\theta_1>\frac{\pi}{2}$ and $\theta_2<\frac{\pi}{2}$ according to the representation in Fig. \ref{fig:scaleInv}a. Results over this difficult set are presented in Fig. \ref{fig:results_reconst} (right). This time, the mean curve method obtains an AUC score of 0.621 and is significantly more accurate than Euler Spiral, with an AUC of 0.492. Clearly, the mean curve reconstruction is far more veridical in such cases.

\section{Conclusions and Future Work}
\label{sec:conclusion}

We have suggested a method for the reconstruction of missing or occluded curves by employing the global statistics of natural contours. At the foundation of our approach is the generation of the mean curve, estimated as the mean of the distribution of natural image curves along their arc length. As we have shown, the estimated mean curve provides an intuitive, visually pleasing, and most importantly, a more veridical reconstruction. A main disadvantage of this procedure, namely the required number of curve fragments sampled to represent the statistics of natural curves loyally, was overcome by exploring and employing certain properties of curves. We have shown that the mean curve is invariant to scale in many cases and that employing this property for sharing fragment samples across scales provides better reconstructions. We also explored midpoint extensibility of the mean curve and have shown that incorporating this property improves the reconstructed curves when the statistical power of the prior is particularly low. 

The data in Fig.~\ref{fig:summary}d shows that for the inducer configuration exemplified, the observed curve centers lie in a small region relative to the distance between the two inducers, exhibiting relatively small variance. In such cases, the center of the mean curve is therefore close to most data points (i.e., real natural curves), and so it could be considered a correct reconstruction (\ie~with low enough error).  Unfortunately, however, this is not always the case, as the mean variance grows larger for certain inducer configurations, as already shown in  Fig. \ref{fig:scaleInv}b (e.g., for inducers facing away from each other). In such cases, where the variance of natural curves is considerable, no single curve can be close to most real curves, {\em regardless of the reconstruction method that generated it}, and thus cannot be considered correct in most cases. 

An immediate consequence of the above is that to obtain correct reconstruction one must reduce the variance of observed data with which reconstruction is attempted. Since the prior in the natural world is fixed, the only way of achieving reduced variance is by introducing additional constraints on the curve fragments that are relevant for a reconstruction query, i.e., by requiring more information about the inducers or the reconstruction scenario.  Such information may include the curvature of the inducers, the shape of the occluder, etc, and incorporating such considerations is part of our ongoing and future research.

\section*{Acknowledgements}
This research was supported in part by Israel Science Foundation (ISF
FIRST/BIKURA Grant 281/15). We also thank the Frankel
Fund and the Helmsley Charitable Trust through the ABC
Robotics Initiative, both at Ben-Gurion University of the
Negev, for their generous support.


{\small
	\bibliographystyle{ieee}
	\bibliography{journal_names,all,ehud_cvpr2018}

\begin{thebibliography}{10}\itemsep=-1pt

\bibitem{Ben-Shahar_Ben-Yosef_2015_PAMI}
O.~Ben-Shahar and G.~Ben-Yosef.
\newblock Tangent bundle elastica and computer vision.
\newblock {\em IEEE Trans. Pattern Anal. Mach. Intell.}, 37:161--174, 2015.

\bibitem{Ben-Yosef_Ben-Shahar_2012_PAMI}
G.~Ben-Yosef and O.~Ben-Shahar.
\newblock A tangent bundle theory for visual curve completion.
\newblock {\em IEEE Trans. Pattern Anal. Mach. Intell.}, 34(7):1263--1280,
  2012.

\bibitem{Bruckstein_Netravali_1990_CVGIP}
A.~Bruckstein and A.~Netravali.
\newblock On minimal energy trajectories.
\newblock {\em CVGIP}, 49(3):283--296, 1990.

\bibitem{Cootes_etal_1995_CVIU}
T.~F. Cootes, C.~J. Taylor, D.~H. Cooper, and J.~Graham.
\newblock Active shape models-their training and application.
\newblock {\em CVIU}, 61:38--59, 1995.

\bibitem{eiter_and_mannila_1994_frechet}
T.~Eiter and H.~Mannila.
\newblock Computing discrete fr{\'e}chet distance.
\newblock Technical report, Tech. Report CD-TR 94/64, Information Systems
  Department, Technical University of Vienna, 1994.

\bibitem{felzenszwalb_and_McAllester_2006_CVPRW}
P.~Felzenszwalb and D.~McAllester.
\newblock A min-cover approach for finding salient curves.
\newblock In {\em Computer Vision and Pattern Recognition Workshop, 2006.
  CVPRW'06. Conference on}, pages 185--185. IEEE, 2006.

\bibitem{Geisler_etal_2001_VR}
W.~Geisler, J.~Perry, B.~Super, and D.~Gallogly.
\newblock Edge co-occurrence in natural images predicts contour grouping
  performance.
\newblock {\em Vision Res.}, 41(6):711--724, 2001.

\bibitem{guo_kimia_2012_CVPRW}
Y.~Guo and B.~Kimia.
\newblock On evaluating methods for recovering image curve fragments.
\newblock In {\em Computer Vision and Pattern Recognition Workshops (CVPRW),
  2012 IEEE Computer Society Conference on}, pages 9--16. IEEE, 2012.

\bibitem{harary_and_tal_2011_CGF}
G.~Harary and A.~Tal.
\newblock The natural 3d spiral.
\newblock In {\em Computer Graphics Forum}, volume~30, pages 237--246. Wiley
  Online Library, 2011.

\bibitem{harary_tal_2012_CompGeom}
G.~Harary and A.~Tal.
\newblock 3d euler spirals for 3d curve completion.
\newblock {\em Computational Geometry}, 45(3):115--126, 2012.

\bibitem{Horn_1983_ATMS}
B.~Horn.
\newblock The curve of least energy.
\newblock {\em ACM Trans. Math. Software}, 9(4):441--460, 1983.

\bibitem{Kellman_Shipley_1991_Cognitive_Psychology}
P.~Kellman and T.~Shipley.
\newblock A theory of visual interpolation in object perception.
\newblock {\em Cognitive Psychology}, 23:141--221, 1991.

\bibitem{Kimia_etal_1999_POCV}
B.~Kimia, L.~Frankel, and A.~Popescu.
\newblock Euler spiral for shape completion.
\newblock In {\em Proc. POCV}, 1999.

\bibitem{Lawlor_Zucker_2013_NIPS}
M.~Lawlor and S.~W. Zucker.
\newblock Third-order edge statistics: Contour continuation, curvature, and
  cortical connections.
\newblock In {\em NIPS}, 2013.

\bibitem{Martin_etal_2001_ICCV}
D.~Martin, C.~Fowlkes, D.~Tal, and J.~Malik.
\newblock A database of human segmented natural images and its application to
  evaluating segmentation algorithms and measuring ecological statistics.
\newblock In {\em ICCV}, pages 416--423, 2001.

\bibitem{mio_etal_2006_IJCV}
W.~Mio, A.~Srivastava, and X.~Liu.
\newblock Contour inferences for image understanding.
\newblock {\em Int. J. Comput. Vision}, 69(1):137--144, 2006.

\bibitem{mohri_etal_2012_foundationsOfML}
M.~Mohri, A.~Rostamizadeh, and A.~Talwalkar.
\newblock {\em Foundations of machine learning}.
\newblock MIT press, 2012.

\bibitem{Mumford_1994_in_Algebric_Geometry_and_its_applications}
D.~Mumford.
\newblock Elastica and computer vision.
\newblock In B.~Chandrajit, editor, {\em Algebric Geometry and its
  applications}. Springer-Verlag, 1994.

\bibitem{ren_malik_2002_ECCV}
X.~Ren and J.~Malik.
\newblock A probabilistic multi-scale model for contour completion based on
  image statistics.
\newblock {\em ECCV}, pages 312--327, 2002.

\bibitem{Ruderman_Bialek_1994_PRLetter}
D.~Ruderman and W.~Bialek.
\newblock Statistics of natural images: Scaling in the woods.
\newblock {\em Phys. Rev. Lett.}, 73(6):814--818, 1994.

\bibitem{sebastian_etal_2000_curve_atlas}
T.~B. Sebastian, J.~J. Crisco, P.~N. Klein, and B.~B. Kimia.
\newblock Constructing 2d curve atlases.
\newblock In {\em Mathematical Methods in Biomedical Image Analysis, 2000.
  Proceedings. IEEE Workshop on}, pages 70--77. IEEE, 2000.

\bibitem{Sharon_etal_2000_PAMI}
E.~Sharon, A.~Brandt, and R.~Basri.
\newblock Completion energies and scale.
\newblock {\em IEEE Trans. Pattern Anal. Mach. Intell.}, 22(10):1117--1131,
  2000.

\bibitem{shibata_etal_2011_ACCV}
T.~Shibata, A.~Iketani, and S.~Senda.
\newblock Image inpainting based on probabilistic structure estimation.
\newblock {\em Proc. ACCV}, pages 109--120, 2011.

\bibitem{sun_etal_2005_tog}
J.~Sun, L.~Yuan, J.~Jia, and H.-Y. Shum.
\newblock Image completion with structure propagation.
\newblock {\em ACM Transactions on Graphics (ToG)}, 24(3):861--868, 2005.

\bibitem{Ullman_1976_BC}
S.~Ullman.
\newblock Filling-in the gaps: The shape of subjective contours and a model for
  their creation.
\newblock {\em Biol. Cybern.}, 25(1):1--6, 1976.

\bibitem{voronin_etal_2012_SPIE}
V.~Voronin, V.~Marchuk, A.~Sherstobitov, and K.~Egiazarian.
\newblock Image inpainting using cubic spline-based edge reconstruction.
\newblock In {\em IS\&T/SPIE Electronic Imaging}, pages 82950I--82950I.
  International Society for Optics and Photonics, 2012.

\bibitem{walton_meek_2008_CRV}
D.~J. Walton and D.~S. Meek.
\newblock An improved euler spiral algorithm for shape completion.
\newblock In {\em Computer and Robot Vision, 2008. CRV'08. Canadian Conference
  on}, pages 237--244. IEEE, 2008.

\bibitem{walton_etal_2009_JCAM}
D.~J. Walton and D.~S. Meek.
\newblock G1 interpolation with a single cornu spiral segment.
\newblock {\em Journal of Computational and Applied Mathematics},
  223(1):86--96, 2009.

\bibitem{wei_liu_2016_SigImgVideoProc}
Y.~Wei and S.~Liu.
\newblock Domain-based structure-aware image inpainting.
\newblock {\em Signal, Image and Video Processing}, 10(5):911--919, 2016.

\bibitem{Weiss_1988_CVGIP}
I.~Weiss.
\newblock 3d shape representation by contours.
\newblock {\em CVGIP}, 41:80--100, 1988.

\bibitem{Williams_Jacobs_1997_NC}
L.~Williams and D.~Jacobs.
\newblock Stochastic completion fields: A neural model of illusory contour
  shape and salience.
\newblock {\em Neural Comp.}, 9(4):837--858, 1997.

\bibitem{xu_etal_2013_CVPR}
J.~Xu, M.~D. Collins, and V.~Singh.
\newblock Incorporating topological constraints within interactive segmentation
  and contour completion via discrete calculus.
\newblock In {\em CVPR}, 2013.

\bibitem{zhou_etal_2012_CompsAndGraphics}
H.~Zhou, J.~Zheng, and X.~Yang.
\newblock Euler arc splines for curve completion.
\newblock {\em Computers \& Graphics}, 36(6):642--650, 2012.

\end{thebibliography}
}

\end{document}